\documentclass[10pt,twocolumn,letterpaper]{article}

\usepackage{cvpr}              % To produce the CAMERA-READY version
\usepackage{wrapfig}
\usepackage{subcaption}
\renewcommand{\paragraph}[1]{\vspace{.5em}\noindent\textbf{#1.}}

\definecolor{cvprblue}{rgb}{0.21,0.49,0.74}
\usepackage[pagebackref,breaklinks,colorlinks,allcolors=cvprblue]{hyperref}

\def\paperID{} % *** Enter the Paper ID here
\def\confName{CVPR}
\def\confYear{2026}

\title{GATS: Gaussian Aware Temporal Scaling Transformer for Invariant 4D Spatio-Temporal Point Cloud Representation}

\author{Jiayi Tian\\
State Key Laboratory of Human-Machine Hybrid Augmented Intelligence \\
Institute of Artificial Intelligence and Robotics, Xi'an Jiaotong University\\
{\tt\small tianreg@stu.xjtu.edu.cn}
\and
Jiaze Wang\\
Harbin Institute of Technology, Shenzhen, Pengcheng Laboratory\\
{\tt\small 24b951071@stu.hit.edu.cn}
}

\begin{document}
\maketitle
\begin{abstract}
Understanding 4D point cloud videos is essential for enabling intelligent agents to perceive dynamic environments. However, temporal scale bias across varying frame rates and distributional uncertainty in irregular point clouds make it highly challenging to design a unified and robust 4D backbone. Existing CNN or Transformer based methods are constrained either by limited receptive fields or by quadratic computational complexity, while neglecting these implicit distortions. To address this problem, we propose a novel dual invariant framework, termed \textbf{Gaussian Aware Temporal Scaling (GATS)}, which explicitly resolves both distributional inconsistencies and temporal. The proposed \emph{Uncertainty Guided Gaussian Convolution (UGGC)} incorporates local Gaussian statistics and uncertainty aware gating into point convolution, thereby achieving robust neighborhood aggregation under density variation, noise, and occlusion. In parallel, the \emph{Temporal Scaling Attention (TSA)} introduces a learnable scaling factor to normalize temporal distances, ensuring frame partition invariance and consistent velocity estimation across different frame rates. These two modules are complementary: temporal scaling normalizes time intervals prior to Gaussian estimation, while Gaussian modeling enhances robustness to irregular distributions. Our experiments on mainstream benchmarks MSR-Action3D (\textbf{+6.62\%} accuracy), NTU RGBD (\textbf{+1.4\%} accuracy), and Synthia4D (\textbf{+1.8\%} mIoU) demonstrate significant performance gains, offering a more efficient and principled paradigm for invariant 4D point cloud video understanding with superior accuracy, robustness, and scalability compared to Transformer based counterparts.
\end{abstract}
    
\section{Introduction}
\label{sec:intro}

4D point cloud videos, which extend 3D space with 1D time, provide a natural representation of dynamic physical environments by capturing both geometric structures and temporal motions~\cite{wang2019dynamic}. They are increasingly important for enabling intelligent agents to perceive dynamics, understand environmental changes, and interact effectively with the world~\cite{zhong2022no}. Consequently, point cloud video modeling has attracted growing attention~\cite{wiesmann2022retriever, qian2022pointnext, guo2016multi}, with applications spanning robotic~\cite{saha2022translating}, AR/VR~\cite{su2015multi}, and SLAM systems~\cite{qi2016volumetric, fan2019pointrnn}.

\begin{figure*}[ht]
    \centering
    \includegraphics[width=\linewidth]{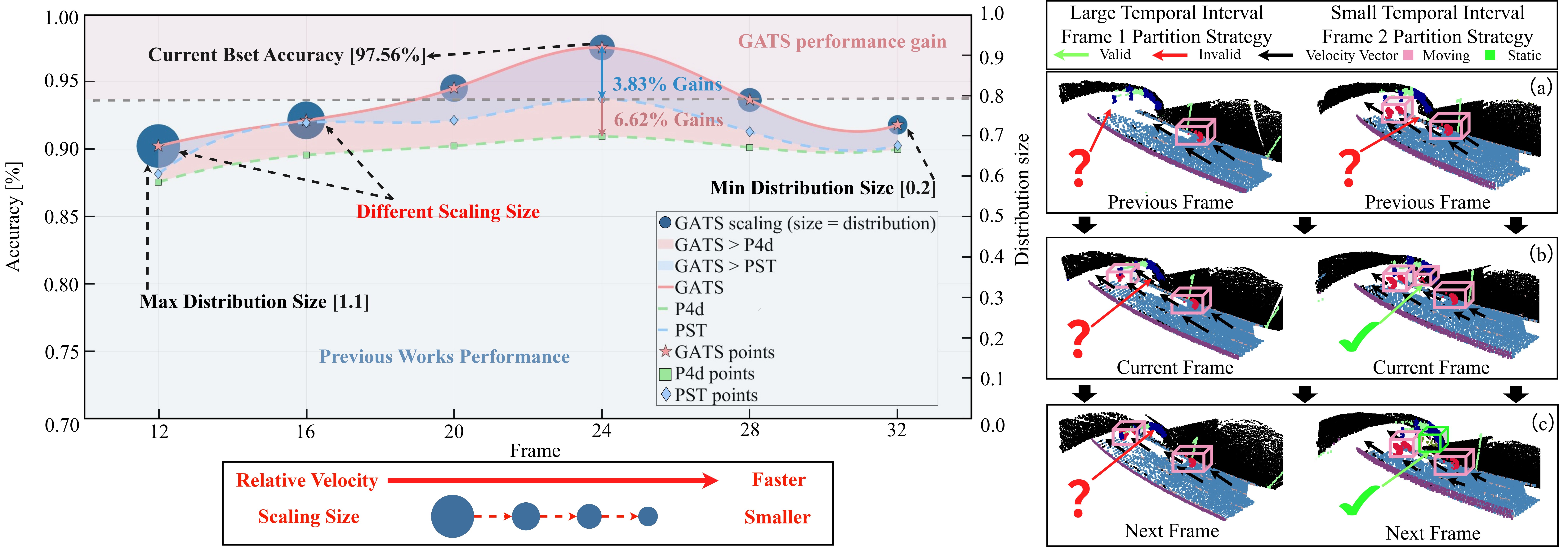}
    \caption{Motivation. \textbf{Right:} The right panel illustrates video sequences under different frame rate partitions: with a large temporal interval, some moving objects (second row) and static objects (third column) disappear, while they remain visible with a small temporal interval. This indicates that varying frame rate partitions may cause certain velocity features to vanish. \textbf{Left: }The left panel illustrates that GATS can adaptively adjust the scaling distribution across different frame rate partitions, thereby effectively mitigating the relative velocity bias introduced by frame rate variations and reducing fluctuations in accuracy. Consequently, GATS achieves improvements in ACC of 6.62\% and 3.83\% over P4D and PST, respectively.}
    \label{fig1}
\end{figure*}

Despite the remarkable progress in static 3D point cloud understanding~\cite{wiesmann2022retriever, saha2022translating, guo2016multi}, building effective backbones for dynamic 4D point cloud sequences remains a significant challenge~\cite{liu2019flownet3d}. Unlike conventional RGB videos with structured grids~\cite{liu2024difflow3d}, point cloud videos are inherently irregular and unordered in space, making grid based methods such as 3D convolutions~\cite{wang2022pointmotionnet} unsuitable. A straightforward approach is voxelization~\cite{wen2022point}, which converts raw 4D sequences into structured grids for 4D convolutions. However, voxelization inevitably introduces quantization errors~\cite{liu2023leaf} and suffers from inefficiency due to the sparsity and scale of 4D data~\cite{zhong2022no}. Another line of research~\cite{gu2022efficiently, gu2020hippo, zhu2024vision} directly processes raw 4D sequences using convolutional networks~\cite{liu2024vmamba} or transformers~\cite{fan2021point, shen2023pointcmp}. Yet, CNNs are limited by their local receptive fields~\cite{qi2017pointnet++}, while transformers incur quadratic complexity~\cite{wang2022pointmotionnet}.

Beyond efficiency and scalability, we identify two fundamental distortions in point cloud video modeling—complementary facets of a unified spatio-temporal misrepresentation (\cref{fig1}): 
(1) \textbf{Distributional uncertainty}: current geometric convolutions only consider Euclidean distances, ignoring local distributional shape and uncertainty. Dynamic point clouds, however, naturally exhibit density variation, noise, occlusion, and missing points, which degrade robustness.
(2) \textbf{Temporal scale bias}: Under different frame intervals, the same physical motion may be discretized into different relative velocity estimates, leading to inconsistencies and distortions in spatio-temporal representations. Existing methods typically rely on fixed frame partitioning or sampling rates. 

Motivated by these observations, we propose \textbf{Gaussian Aware Temporal Scaling (GATS)}, a dual-invariant Transformer framework for 4D point cloud video modeling. The central insight is that a collaborative calibration mechanism is required to jointly normalize geometric distributions and temporal motions, thereby achieving truly invariant and robust spatio-temporal representations. Specifically, we introduce two complementary modules:
(1) \textbf{Uncertainty Guided Gaussian Convolution (UGGC)} that augments geometric kernels with Gaussian statistics (mean, covariance, and uncertainty indicators), enabling robust neighborhood aggregation under density variation and noise. different frame rates. 
(2) \textbf{Temporal Scaling Attention (TSA)}: with a learnable normalization factor, it enables velocity-invariant temporal modeling, ensures consistent frame partitioning, and reduces distortions across frame rates.

\begin{figure*}[t]
    \centering
    \includegraphics[width=\linewidth]{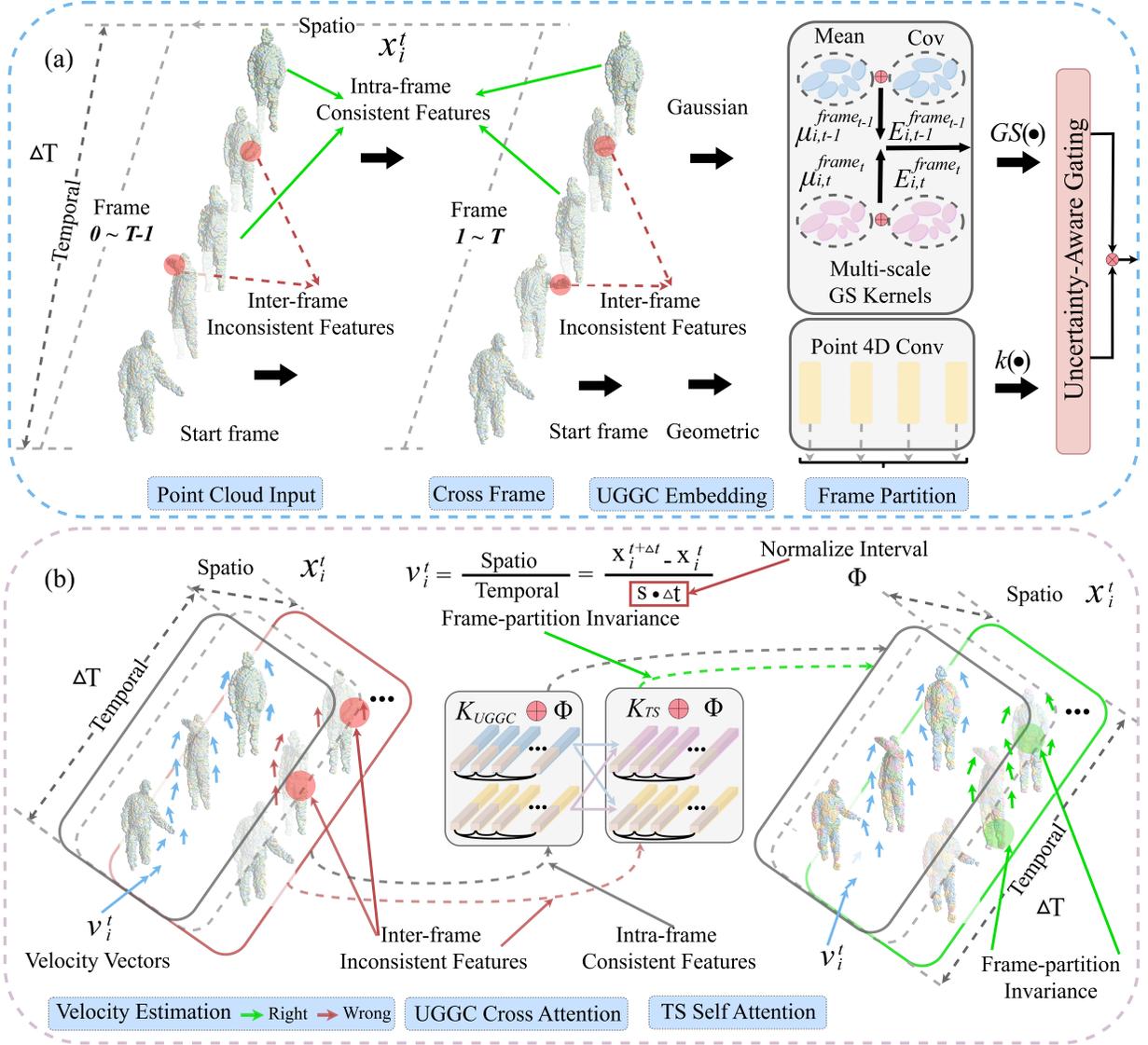}
    \caption{Pipeline. The overall network backbone consists of two core modules: \textbf{(a) UGGC Module.} After the point cloud is fed into the network, the spatial variations of $x_i^t$ generate cross frame representations. However, different cross frames often lead to inter frame inconsistencies. The UGGC module extracts local Gaussian features and incorporates an uncertainty aware gating mechanism to jointly model geometric and Gaussian local features of 4D point clouds, thereby enhancing the robustness of feature extraction. \textbf{(b) TSA Module.} Under different frame rates, the estimation of relative velocity $_i^t$ varies, and as the temporal dimension progresses, motion features tend to produce inter frame inconsistencies. To address this, the TSA module introduces a learnable scaling factor $s$ to normalize temporal distances, achieving frame partition invariance and ensuring consistent relative velocity estimation across varying frame rates.
}
    \label{fig2}
\end{figure*}

These two modules are naturally synergistic: temporal scaling normalizes time intervals before Gaussian estimation, preventing variance inflation across different frame rates, while Gaussian modeling provides distributional robustness for spatio-temporal neighborhoods. The pipeline is shown in \cref{fig2}. To the best of our knowledge, this is the \textbf{first work} to explicitly introduce relative velocity estimation in spatio-temporal point cloud modeling.

Overall, our main contributions are as follows:  
\begin{itemize}
    \item We propose a novel 4D backbone, \textbf{GATS}, that explicitly addresses two implicit distortions in point cloud video modeling: temporal scale bias and distributional uncertainty.  
    \item We introduce an \textbf{UGGC module} that incorporates local Gaussian statistics and uncertainty aware gating into P4DConv, enhancing robustness to noise, occlusion, and density variation.  
    \item We design a \textbf{TSA} module that achieves frame partition invariance by rescaling temporal metrics, improving consistency across varying frame rates and sampling strategies.  
    \item Extensive experiments on multiple 4D understanding benchmarks demonstrate that GATS achieves superior performance and robustness compared to baseline methods, while maintaining high efficiency and scalability.  
\end{itemize}
\section{Related Work}
\label{sec:related}
\subsection{Backbones for 4D point cloud videos: CNNs, Transformers, and SSMs}
Backbones for dynamic 4D point cloud sequences have evolved along three major paradigms: CNNs, Transformers, and State Space Models (SSMs)~\cite{fan2019pointrnn, fan2021point, liang2024pointmamba}. 
Early CNN based methods either convert raw 4D sequences into structured grids (\eg, voxels or BEV) to apply 3D/4D convolutions, or directly operate on raw points with spatio-temporal kernels~\cite{fan2021pstnet}. While grid conversion enables standard convolutions, it introduces quantization artifacts and loses geometric fidelity; point based CNNs are constrained by locally restrained receptive fields and limited long range temporal modeling.
Transformer based backbones~\cite{fan2021point} enlarge receptive fields using global self-attention and often pair local point convolutions with temporal attention across timestamps; efficiency oriented variants employ compact mid-level tokens or self-supervised training to mitigate annotation cost. 
However, quadratic complexity and sensitivity to frame partitioning remain bottlenecks, leading to high overhead and temporal scale bias under varying sampling strategies.
Recently, SSM based designs (\eg, Mamba style modules~\cite{ma2024u, xing2024segmamba}) achieve linear complexity for long sequence modeling via parallel scans and data dependent state transitions, offering scalable temporal coverage with global receptive fields.
Distinct from prior art, our framework emphasizes \emph{dual invariance}: it mitigates temporal scale bias through temporal scaling attention and enhances distributional robustness via Gaussian aware convolution, complementing architectural efficiency with invariance aware modeling.

\subsection{Geometric modeling on dynamic 4D point clouds}
Research on geometric modeling for dynamic point clouds follows two threads: (i) \emph{dynamic as signal} methods that take temporal variation itself as the modeling target~\cite{liu2023contrastive, qi2017pointnet}, and (ii) \emph{geometry only} methods that rely on static descriptors without explicit temporal reasoning~\cite{jing2024x4d}.
Dynamic as signal approaches explicitly track points or estimate scene flow to align correspondences across frames, enabling motion aware grouping and temporal aggregation (\eg, recurrent point modules and PointNet++ extensions with temporal grouping or flow based alignment~\cite{qian2022pointnext, qi2017pointnet++}). 
While explicit motion preserves identities, performance is sensitive to tracking errors, motion inconsistency, and point entry/exit in sparse or noisy scenes. 
Subsequent methods (\eg, PST style spatio temporal convolutions~\cite{fan2019pointrnn, fan2021point, liang2024pointmamba}) construct 4D neighborhoods to learn motion patterns implicitly without hard associations, improving robustness but remaining limited by local receptive fields.
Geometry only approaches emphasize robust per frame spatial descriptors and aggregate them across time, avoiding explicit tracking or flow. 
Typical designs use Euclidean kernels, anisotropic neighborhood selection, and hierarchical local feature extraction~\cite{huang2022unsupervised}; hybrid voxel–point pipelines alleviate sparsity and efficiency issues but still lack explicit statistical modeling.
To address these limitations, distribution aware schemes introduce statistical cues (\eg, Gaussian weighting, covariance guided anisotropy) into neighborhood aggregation. 
Our method advances this direction with \emph{multi-scale Gaussian weighting} for heterogeneous densities and \emph{uncertainty aware gating} via covariance conditioning, yielding distributionally robust features without explicit tracking.

\subsection{Spatio-temporal decoupled representations for 4D point cloud videos}
Decoupling space and time is a pragmatic strategy for 4D point clouds: spatial sampling is irregular and unordered, whereas temporal sampling is typically regular~\cite{fan2022point}. 
Prior frameworks commonly adopt a two branch design that combines \emph{intra frame spatial encoding} for local geometric patterns with \emph{inter frame temporal modeling} for long range dependencies. 
Intra frame modules include point based CNNs and local operators (\eg, PointNet++ variants and PST style kernels~\cite{shen2023masked, fan2021point, qi2017pointnet++}) that build robust neighborhoods, perform hierarchical aggregation, and leverage anisotropic receptive fields; hybrid voxel–point tokens provide compact intermediate representations but introduce quantization artifacts. 
Inter frame modules range from RNNs and temporal pooling to global attention and SSMs: Transformers (\eg, temporal attention over all timestamps~\cite{liu2025mamba4d, liang2024pointmamba}) capture global semantics at quadratic cost, while Mamba style SSMs provide linear time long sequence coverage via data dependent state transitions and parallel scans.
Despite the effectiveness of decoupling, most methods assume fixed frame rates or uniform partitions, which induces \emph{(1) temporal scale bias}: identical physical motion maps to different discrete temporal distances and velocities under varying sampling. 
Moreover, spatial neighborhoods are commonly treated with purely geometric kernels, overlooking \emph{(2) distributional uncertainty} (density variation, anisotropy, noise, and missing points).

We retain the decoupling principle and introduce two invariance oriented components: \emph{(1) UGGC module}, which encodes local shape statistics (mean and covariance) together with uncertainty within spatial neighborhoods; \emph{(2) TSA module}, which normalizes temporal metrics and ensures frame partition invariance. This design mitigates temporal scale bias, promotes consistent representations across varying frame rates, and enhances the stability of spatio-temporal modeling. Together, these modules yield decoupled features that remain consistent across frame rates and robust to heterogeneous local distributions, while integrating seamlessly with CNN, Transformer, and SSM backbones.

\section{Methodology}
\label{sec: Method}
\subsection{Problem Definition}
\label{sec:problem}

We consider a point cloud video sequence $\mathcal{P}$ as follows:

\begin{equation}
\mathcal{P} = \{P_t\}_{t=1}^T, \quad P_t = \{(x_i^t, Tf_i^t)\}_{i=1}^{N_t},
\end{equation}

where $\quad P_t$ is the point clouds of each video frame, $x_i^t \in \mathbb{R}^3$ denotes the 3D coordinate of the $i$-th point at frame $t$, $Tf_i^t \in \mathbb{R}^d$ indicates the associated temporal feature, and $N_t$ is the number of points in frame $t$. 
The goal is to learn a spatio-temporal modeling function $\mathcal{F}$ that maps $\{P_t\}_{t=1}^T$ to a target output $y$.

\begin{equation}
\mathcal{F}: \{P_t\}_{t=1}^T \mapsto y,
\end{equation}

where $y$ can be an action label, per point segmentation mask, or vedio modeling.

Compared with conventional videos, point cloud videos face two key distortions: (i) \textbf{Distributional uncertainty} from irregular geometry and noise. (ii) \textbf{Temporal scale bias} from inconsistent relative velocity estimates under varying frame rates.

Moreover, uncalibrated geometric representations distort motion perception, while unnormalized motion interferes with geometric understanding. This dual dilemma must be resolved through a complementary fusion framework. To address these issues, we propose \textbf{Gaussian Aware Temporal Scaling (GATS)}, a dual-invariant framework that introduces (2) \emph{Uncertainty Guided Gaussian Convolution} for distributional robustness, and (2) \emph{Temporal Scaling Attention} for frame partition invariance.

% ------------------------------
\subsection{Uncertainty Guided Gaussian Convolution}
\label{sec:gaussian}
 
Geometric convolutions typically rely on Euclidean distance, ignoring local distributional shape and uncertainty. 
Dynamic point clouds, however, exhibit density variation, noise, and occlusion. 
We propose to incorporate Gaussian statistics into 4D Points convolution to enhance robustness.

\paragraph{Local Gaussian Estimation}  
For a center point $x_i^t$, its 4D neighborhood $N(i,t)$ is modeled by mean and covariance:
\begin{equation}
\mu_{i,t} = \frac{1}{|N(i,t)|} \sum_{x_i^t \in N(i,t)} x_i^t, \quad
\end{equation}
\begin{equation}
\Sigma_{i,t} = \frac{1}{|N(i,t)|} \sum_{x_i^t\in N(i,t)} (x_i^t-\mu_{i,t})(x_i^t-\mu_{i,t})^\top.
\end{equation}
Here $\mu_{i,t} \in \mathbb{R}^3$ is the local centroid, and $\Sigma_{i,t} \in \mathbb{R}^{3 \times 3}$ encodes distributional anisotropy.

\paragraph{Gaussian Weighted Convolution}  
Building upon this estimation, we design a Gaussian weighted convolution that integrates both geometric kernels $k(\cdot)$ and Gaussian statistical $GS(\cdot)$ likelihoods. The aggregation weight is defined as:
\begin{equation}
w(x) = k(x_i^t-\mu_{i,t}) \cdot \exp\!\left(-\tfrac{1}{2}(x_i^t-\mu_{i,t})^\top \Sigma_{i,t}^{-1} (x_i^t-\mu_{i,t})\right).
\end{equation}
To further adapt to heterogeneous densities, we employ multi-scale Gaussian kernels with $\sigma \in \{0.5r, r, 3r\}$.

\paragraph{Uncertainty Aware Gating} 
While Gaussian weighting improves robustness, the reliability of local statistics may still vary under severe noise or occlusion. To adaptively balance standard and robust features, we introduce an uncertainty aware gating mechanism. Using the condition number $\text{cond}(\Sigma_{i,t})$ or its eigenvalue spectrum as an uncertainty indicator, we define:
\begin{equation}
f' = \alpha f + (1-\alpha) f_{\text{robust}}, \quad 
\alpha = \phi(\text{cond}(\Sigma_{i,t})),
\end{equation}
where $f$ is the standard convolution feature, $f_{\text{robust}}$ is a complementary branch (e.g., larger receptive field), and $\phi(\cdot)$ maps uncertainty to $[0,1]$. This gating ensures that the model adaptively emphasizes robust features when uncertainty is high, while preserving efficiency in stable regions.

\subsection{Temporal Scaling Attention}
\label{sec:scaling}
  
Existing Transformer based models often rely on discrete frame indices $|t-t'|$ as temporal bias. However, the same physical motion may correspond to different discrete intervals $\Delta t$ under varying frame rates, leading to inconsistent velocity estimation. To address this temporal scale bias, we introduce a temporal scaling factor that normalizes such discrepancies and ensures consistent motion representation.

\paragraph{Relative Velocity}  
First, given a point $x_i^t \in \mathbb{R}^3$ at time $t$, the relative velocity $v_i^t \in V_i^t$is defined as:
\begin{equation}
v_i^t = \frac{x_i^{t+\Delta t} - x_i^t}{\Delta t},
\end{equation}
where $\Delta t\in \Delta T$ denotes the frame interval. To eliminate the dependency on arbitrary frame rates, we introduce a learnable or estimable scaling factor $s \in \mathbb{R}^+$:
\begin{equation}
\Delta t' = s \cdot \Delta t, \quad 
v_i^t = \frac{x_i^{t+\Delta t} - x_i^t}{s \cdot \Delta t}.
\end{equation}
This normalization ensures that velocity estimation remains consistent across different temporal partitions.

\paragraph{Temporal Scaling Attention}  
Building upon this formulation, we embed the scaling factor into the attention mechanism. Specifically, the scaled temporal distance modifies the positional bias:
\begin{equation}
\text{Attn}(q_t, k_{t'}) = \frac{q_t k_{t'}^\top}{\sqrt{d}} + \beta \cdot \Phi(s \cdot |t-t'|),
\end{equation}
where $q_t, k_{t'} \in \mathbb{R}^d$ are query and key vectors. In our design, the key vector further integrates two complementary components: $K_{\text{UGGC}}$ from the Uncertainty Guided Gaussian Convolution branch and $K_{\text{TS}}$ from the Temporal Scaling branch; $\Phi(\cdot)$ maps temporal distance to bias, and $\beta$ is a learnable weight. Unlike conventional additive bias, this scaling redefines the temporal metric space, thereby achieving frame partition invariance.

\paragraph{Geometric Feature Extraction}  
The temporal scaling factor also benefits geometric feature extraction. In P4D convolution, the temporal neighborhood radius $r_t$ is rescaled as:
\begin{equation}
r_t' = s \cdot r_t,
\end{equation}
which guarantees consistent neighborhood selection regardless of frame rate variations.

\paragraph{Synergy with Temporal Scaling}  
Finally, temporal scaling naturally complements Gaussian based modeling. By normalizing the temporal radius after Gaussian estimation, it prevents variance inflation across different frame rates and ensures the comparability of Gaussian attributes. This synergy highlights the dual role of temporal scaling: stabilizing motion representation and enhancing distributional robustness.

\section{Experiments}

\subsection{Experimental Setup}

Our experimental validation is performed on three widely recognized benchmarks. For the 3D action recognition task, we utilize MSR-Action3D \citep{li2010action} and NTU RGBD \citep{shahroudy2016ntu}. For 4D semantic segmentation, we employ the Synthia 4D dataset \citep{choy20194d}. A comprehensive description of the experimental dataset can be found in Appendix 1.1. Additionally, further details regarding the experimental settings are presented in Appendix 1.2.

\begin{figure}[ht]
    \centering
    \includegraphics[width=0.95\linewidth]{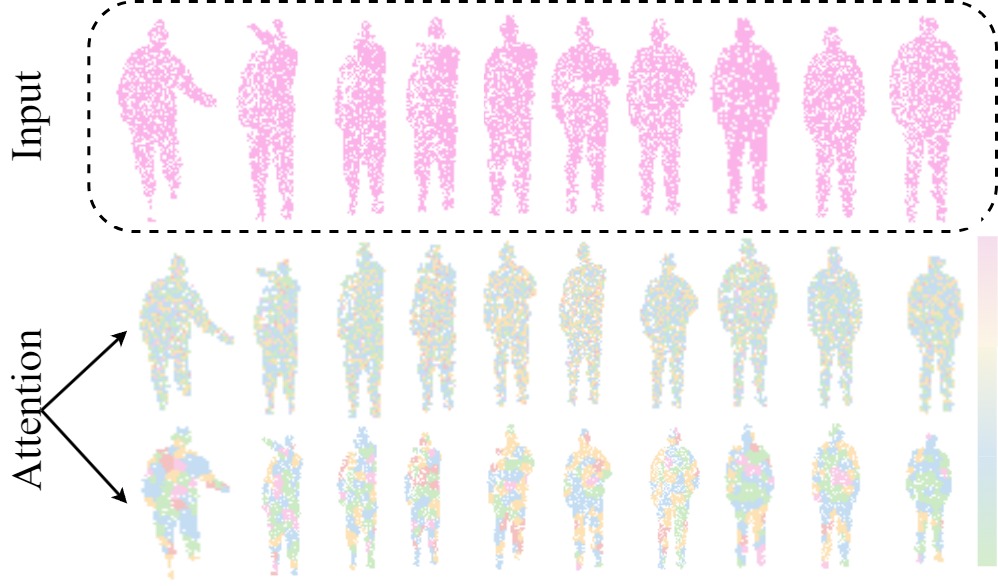}
    \caption{MSR-attention Results. Attention visualization showing focus on key regions and spatio-temporal dynamics.}
    \label{MSR}
\end{figure}

\begin{table}
\caption{Action recognition accuracy comparison (\%) on MSR-Action3D.}
\label{tab3d}
\centering
\resizebox{\linewidth}{!}{

\begin{tabular}{lccc}
\toprule
\textbf{Methods} & \textbf{Input} & \textbf{Frames} & \textbf{Accuracy (\%)} \\
\midrule

Actionlet \cite{wang2012mining} & skeleton & all & 88.21 \\
MeteorNet \cite{liu2019meteornet} & point & 12 & 86.53 \\
PSTNet \cite{fan2021pstnet} & point & 12 & 87.88 \\
P4Transformer \cite{fan2021point} & point & 12 & 87.54 \\
PST-Transformer \cite{fan2022point} & point & 12 & 88.15 \\
GATS (Ours) & point & 12 & \textbf{90.24} \\
Vieira et al. \cite{vieira2012stop} & depth & 20 & 78.20 \\
GATS (Ours) & point & 20 & \textbf{94.54} \\
MeteorNet \cite{liu2019meteornet} & point & 24 & 88.50 \\
PSTNet \cite{fan2021pstnet} & point & 24 & 91.20 \\
P4Transformer \cite{fan2021point} & point & 24 & 90.94 \\
PST-Transformer \cite{fan2022point} & point & 24 & 93.73 \\
PSTNet++\cite{fan2021deep} & point & 24 & 92.68 \\
PPTr+C2P\cite{zhang2023complete} & point & 24 & 94.76 \\
PointCPSC\cite{shen2023pointcmp} & point & 24 & 92.68 \\
MaST-Pre\cite{shen2023masked} & point & 24 & 94.08 \\
SequentialPointNet\cite{li2022real} & point & 24 & 92.64 \\
MAMBA4D\cite{liu2025mamba4d} & point & 24 & 93.38 \\
3DInAction\cite{ben20243dinaction} & skeleton & 24 & 92.23 \\
PvNeXt\cite{wang2025pvnext} & point & 24 & 94.77 \\
GATS (Ours) & point & 24 & \textbf{97.56} \\
\bottomrule
\end{tabular}
}
\end{table}

\begin{table}
\caption{Action recognition accuracy (\%) on NTU RGBD.}
\label{tab:ntu60}
\centering
\begin{tabular}{lccr}
\toprule
\textbf{Method} & \textbf{Input} & \textbf{Acc (\%)}  \\
\midrule
SkeleMotion\cite{caetano2019skelemotion}& skeleton & 69.6 \\
GCA-LSTM\cite{liu2017global} & skeleton & 74.4 \\
AttentionLSTM\cite{liu2017skeleton} & skeleton & 77.1 \\
AGC-LSTM\cite{si2019attention} & skeleton & 89.2 \\
AS-GCN\cite{li2019actional} & skeleton & 86.8 \\
VA-fusion\cite{zhang2019view} & skeleton & 89.4 \\
2s-AGCN\cite{shi2019two} & skeleton & 88.5 \\
DGNN\cite{shi2019skeleton} & skeleton & 89.9 \\
HON4D\cite{oreifej2013hon4d} & depth & 30.6 \\
SNV\cite{yang2014super} & depth & 31.8 \\
HOG$^2$\cite{ohn2013joint} & depth & 32.2 \\
Li et al.\cite{li2018unsupervised} & depth & 68.1 \\
Wang et al.\cite{wang2018depth} & depth & 87.1 \\
MVDI\cite{xiao2019action} & depth & 84.6 \\
PointNet++\cite{qi2017pointnet++} & point & 80.1 \\
3DV\cite{wang20203dv} & voxel & 84.5 \\
3DV-PointNet++\cite{wang20203dv} & voxel+point & 88.8 \\
PSTNet\cite{fan2021pstnet} & point & 90.5\\
P4Transformer\cite{fan2021point} & point & 90.2 \\
PST-Transformer\cite{fan2022point} & point & 91.0 \\
SequentialPointNet\cite{li2022real} & point & 90.3 \\
MaST-Pre\cite{shen2023masked} & point & 90.8 \\
PvNeXt\cite{wang2025pvnext} & point & 89.2 \\
\midrule
GATS(Ours)   & point & \textbf{91.7} \\
\bottomrule
\end{tabular}
\end{table}

\begin{table*}[ht]
\centering
\caption{4D semantic segmentation results (mIoU \%) on the Synthia 4D dataset.}
\label{tabseg}
\resizebox{\linewidth}{!}{
\begin{tabular}{l|c|c|c|cccccccc}
\toprule
Method & Input & Frame & Track & Bldn & Road & Sdwlk & Fence & Vegittn & Pole & - \\
\midrule
3D MinkNet14 & voxel & 1 & - & 89.39 & 97.68 & 69.43 & 86.52 & 98.11 & 97.26 & - \\
4D MinkNet14 & voxel & 3 & - & 90.13 & 98.26 & 73.47 & 87.19 & 99.10 & 97.50 & - \\
\hline
PointNet++ & point & 1 & - & 96.88 & 97.72 & 86.20 & 92.75 & 97.12 & 97.09 & - \\
MeteorNet-m & point & 2 & \checkmark & \textbf{98.22} & 97.79 & 90.98 & 93.18 & 98.31 & 97.45 & - \\
MeteorNet-m & point & 2 & $\times$ & 97.65 & 97.83 & 90.03 & 94.06 & 97.41 & 97.79 & - \\
MeteorNet-l & point & 3 & $\times$ & 98.10 & 97.72 & 88.65 & 94.00 & 97.98 & 97.65 & - \\
P4Transformer & point & 1 & - & 96.76 & 98.23 & 92.11 & 95.23 & 98.62 & 97.77 & - \\
P4Transformer & point & 3 & $\times$ & 96.73 & 98.35 & 94.03 & 95.23 & 98.28 & 98.01 & - \\
PST-Transformer & point & 1 & - & 94.46 & 98.13 & 89.37 & 95.84 & \textbf{99.06} & 98.10 & - \\
PST-Transformer & point & 3 & $\times$ & 96.10 & 98.44 & 94.94 & \textbf{96.58} & 98.98 & 98.10 & - \\
MAMBA4D & point & 3 & $\times$ & 96.16 & 98.58 & 92.80 & 94.95 & 97.08 & 98.24 & - \\
\hline
GATS(Ours) & point & 1 & $\times$ & 97.51 & \textbf{98.58} & \textbf{95.62} & 95.54 & 99.01 & 98.27 & - \\
GATS(Ours) & point & 3 & $\times$ & 97.37 & 98.41 & 94.72 & 95.85 & 97.81 & \textbf{98.31} & - \\
\bottomrule
\end{tabular}
}
\smallskip
\resizebox{\linewidth}{!}{
\begin{tabular}{l|c|c|c|cccccccc}
\toprule
Method & Input & Frame & Track & Car & T. Sign & Pedstrn & Bicycl & Lane & T. Light & mIoU \\
\midrule
3D MinkNet14 & voxel & 1 & - & 93.50 & 79.45 & 92.27 & 0.00 & 44.61 & 66.69 & 76.24 \\
4D MinkNet14 & voxel & 3 & - & 94.01 & 79.04 & \textbf{92.62} & 0.00 & 50.01 & 68.14 & 77.46 \\
\hline
PointNet++ & point & 1 & - & 90.85 & 66.87 & 78.64 & 0.00 & 72.93 & 75.17 & 79.35 \\
MeteorNet-m & point & 2 & \checkmark & 94.30 & 76.35 & 81.05 & 0.00 & 74.09 & 75.92 & 81.47 \\
MeteorNet-m & point & 2 & $\times$ & 94.15 & 82.01 & 79.14 & 0.00 & 72.59 & 77.92 & 81.72 \\
MeteorNet-l & point & 3 & $\times$ & 93.83 & 84.07 & 80.90 & 0.00 & 71.14 & 77.60 & 81.80 \\
P4Transformer & point & 1 & - & 95.46 & 80.75 & 85.48 & 0.00 & 74.28 & 74.22 & 82.41 \\
P4Transformer & point & 3 & $\times$ & 95.60 & 81.54 & 85.18 & 0.00 & 75.95 & 79.07 & 83.16 \\
PST-Transformer & point & 1 & - & \textbf{96.80} & 80.41 & 87.58 & 0.00 & 75.25 & 80.84 & 82.92 \\
PST-Transformer & point & 3 & $\times$ & 96.06 & 82.67 & 87.86 & 0.00 & 76.01 & 81.67 & 83.95 \\
MAMBA4D & point & 3 & $\times$ & 95.75 & 82.03 & 84.57 & 0.00 & \textbf{79.35} & 80.74 & 83.35 \\
\hline
GATS(Ours) & point & 1 & $\times$ & 95.62 & 81.08 & 83.11 & 0.00 & 77.19 & \textbf{83.12} & 83.72 \\
GATS(Ours) & point & 3 & $\times$ & 95.80 & \textbf{84.87} & 87.64 & 0.00 & 76.77 & 82.98 & \textbf{84.21} \\
\bottomrule
\end{tabular}
}
\end{table*}

\begin{figure*}[!t]
	\centering   %rebuttal
	\includegraphics[width=\linewidth]{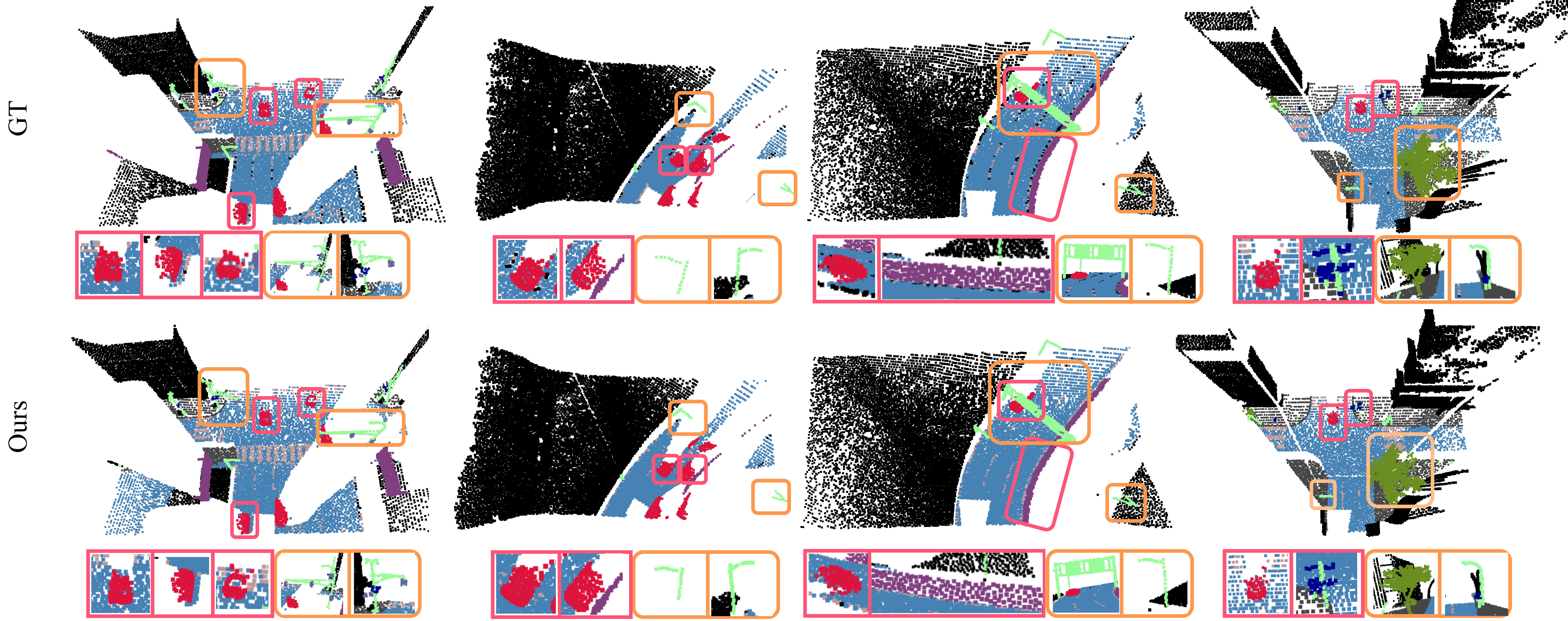}
	\caption{4D Qualitative Results. The rows from top to bottom correspond to the input, GT, and our predictions. Detailed comparative results are highlighted in the enlarged regions of the figure.}
	\label{4D}
\end{figure*}

\subsection{MSR-Action3D}

\paragraph{Quantitative results} \Cref{tab3d} reports the performance of GATS on the MSR-Action3D benchmark. In the 24-frame setting, GATS achieves the highest accuracy of 97.56\%, outperforming the lowest-performing recent model MAMBA4D (93.38\%) by over 4 points. Compared with strong baselines such as PvNeXt (94.77\%) and PST-Transformer (93.73\%), GATS also shows clear improvements. Consistent gains are observed in other settings, reaching 94.54\% at 20 frames and 90.24\% at 12 frames.

\paragraph{Qualitative results} We further visualize attention weights in \cref{MSR}. The results show that the Transformer focuses on relevant regions across frames and highlights key transitional parts of actions. This confirms that our attention mechanism effectively captures spatio-temporal dynamics and can serve as a substitute for explicit point tracking.

\subsection{NTU RGBD}
\paragraph{Quantitative results} The evaluation of action recognition accuracy on the NTU RGB+D dataset is detailed in \cref{tab:ntu60}. Our model, which operates directly on point cloud inputs, achieves a new state-of-the-art accuracy of 91.7\%. This performance surpasses all other listed methods, including strong point based competitors like PST-Transformer (91.0\%) and MaST-Pre (90.8\%). Furthermore, our model outperforms the hybrid voxel+point method, 3DV-PointNet++ (88.8\%), by a significant margin of 2.9\%. This result highlights our framework's superior ability to capture complex spatio-temporal dynamics from raw point sequences, establishing its effectiveness for 3D action recognition.

\subsection{4D Semantic Segmentation}
\paragraph{Quantitative results} As delineated in \cref{tabseg}, our model's 4D semantic segmentation performance was comprehensively benchmarked against state-of-the-art methods on the Synthia 4D dataset. In the challenging multi-frame (frame=3) setting, our model achieves a new state-of-the-art mIoU of 84.21\%, surpassing the previous best, PST-Transformer. Our model also establishes its superiority in the single frame (frame=1) setting, attaining an mIoU of 83.72\% and outperforming all competitors. The significant performance gain from the single frame to the multi frame variant (83.72\% to 84.21\%) effectively demonstrates our architecture's robust ability to leverage temporal information, firmly establishing its overall superiority.

\paragraph{Qualitative results} \Cref{4D} presents a qualitative analysis, comparing our model's predictions against the raw input and GT segmentation. The visualizations confirm that our predictions align closely with the GT across varied and complex scenes. This ability to accurately capture intricate boundaries and fine-grained details underscores the model's high fidelity and strong generalization.

\subsection{Ablation Studies}

To validate the contribution of each key component, we conducted ablation studies on the MSR-Action3D dataset, with results detailed in \cref{tab0}. The full model serves as our baseline, achieving an accuracy of 97.56\%. When the GA module was removed ("w/o GA"), the accuracy experienced a notable drop to 95.12\%. Similarly, excluding the TS module ("w/o TS") reduced the accuracy to 96.16\%. This clear performance degradation upon the removal of either component confirms that both GA and TS are integral to the model's success.

\begin{table}[ht]
\centering
    \caption{Ablation studies on MSR-Action3D.}
    \label{tab0}
    \begin{tabular}{lrr}
        \toprule
        \textbf{Model} & frame & \textbf{Acc(\%)} \\
        \midrule
        (a) Full Model & 24 & 97.56 \\
        (b) w/o UGGC & 24 & 95.12 \\
        (c) w/o TSA & 24 & 96.16 \\
        (d) PSTTransformer (baseline) & 24 & 93.73 \\
        \bottomrule
    \end{tabular}%
\end{table}

\begin{table}[ht]
    \centering
    \caption{Comparison results with advanced architecture processing more frames.}
    \label{tab1}
    \begin{tabular}{llcc}
    \toprule
    \textbf{Method} & \textbf{Backbone} & \textbf{Frame} & \textbf{Acc(\%)} \\
    \midrule
    PvNeXt & CNN & 24 & 94.77 \\
    MAMBA4D & Mamba & 24 & 93.38 \\
    Ours & Transformer & 24 & 97.56 \\
    \bottomrule
    \end{tabular}
\end{table}

\paragraph{Comparative Study on Model Effectiveness and Efficiency} In our comparative analysis (\cref{tab1}), the Transformer model achieved 97.56\% accuracy with 24 frames, outperforming MAMBA4D (32 frames, 93.38\%) and PvNeXt (24 frames, 94.77\%). This result demonstrates not only superior spatio-temporal modeling capability but also greater efficiency. In contrast to models that rely on more frames but yield lower accuracy, our approach achieves superior performance and thus provides a more practical solution for long-sequence tasks.
% \begin{figure}[h]
% \centering
% %\includegraphics[width=3in]{fig5}
% \subfloat[Temporal kernel sizes.]{
% 		\includegraphics[width=0.45\linewidth]{sec/Figure/tem.png}}
% \subfloat[Spatial radius settings.]{
% 		\includegraphics[width=0.45\linewidth]{sec/Figure/spa.png}}
% \caption{Analysis of accuracy with varying temporal and spatial parameters.}
% \label{abl2}
% \end{figure}

\begin{figure}[h]
	\centering   %rebuttal
	\includegraphics[width=\linewidth]{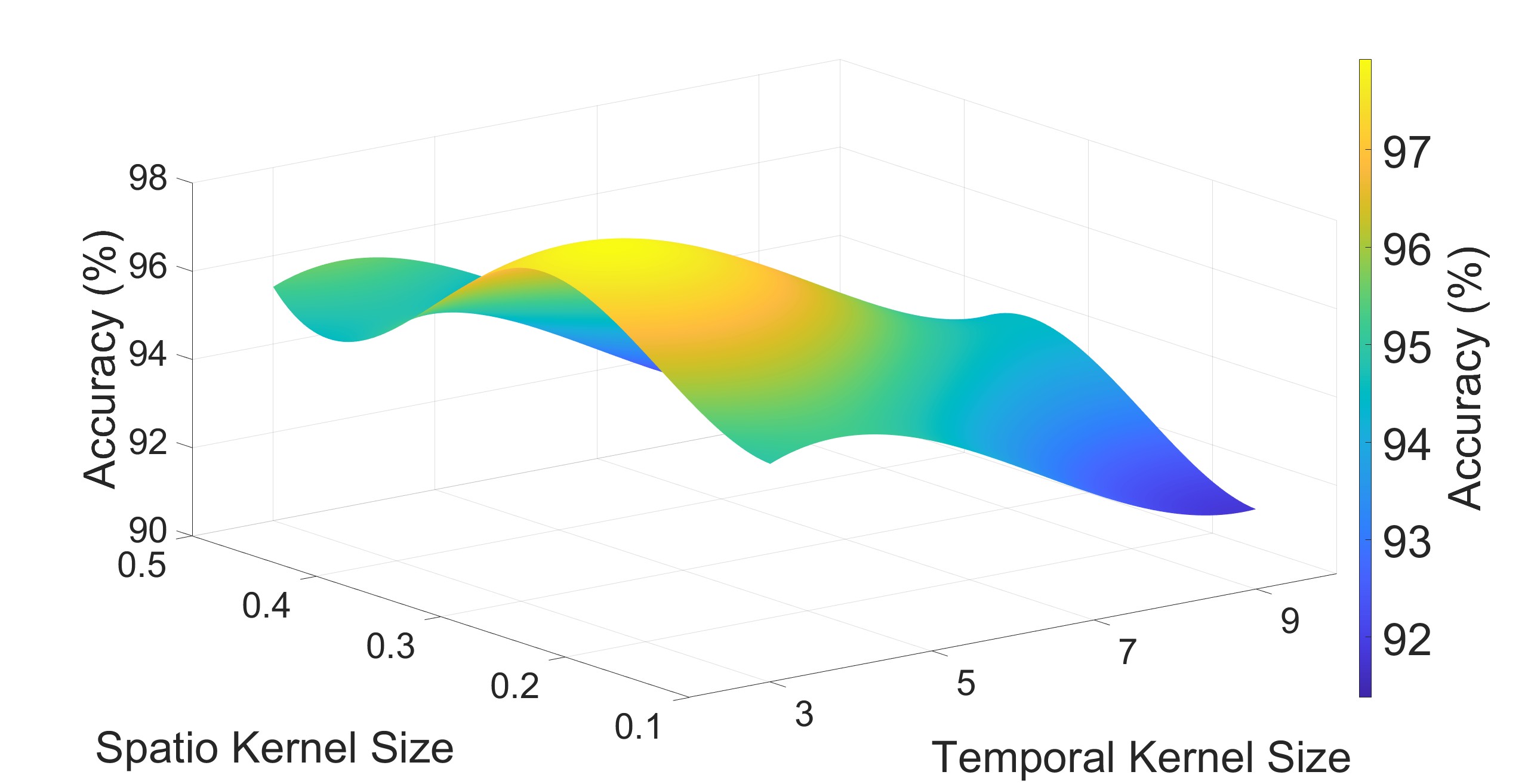}
	\caption{Analysis of accuracy with varying temporal and spatial parameters.}
	\label{abl2}
\end{figure}

\paragraph{Effect of temporal kernel size and spatial radius} \Cref{abl2} illustrates the ablation study on temporal kernel size and spatial radius. The analysis reveals that smaller temporal kernels are advantageous, as the model achieves its peak accuracy at the smallest tested size of 3, after which performance monotonically decreases with larger kernels. In contrast, the spatial radius requires an optimal balance; accuracy improves from a radius of 0.1, peaks at 97.56\% with a radius of 0.3, and then distinctly degrades at larger values. 

% \begin{wrapfigure}{t}{\linewidth}
%     \centering
%     \vspace{-15pt} % 可选：微调垂直间距
%     \begin{minipage}{\linewidth}
%         \centering
%         \begin{subfigure}{\linewidth}
%             \centering
%             \includegraphics[width=\linewidth]{sec/Figure/tem.png}
%             \caption{Temporal kernel sizes}
%             \label{fig:side_a}
%         \end{subfigure}

%         \vspace{1em} 

%         \begin{subfigure}{\linewidth}
%             \centering
%             \includegraphics[width=\linewidth]{sec/Figure/spa.png}
%             \caption{Spatial radius settings}
%             \label{fig:side_b}
%         \end{subfigure}
        
%         % 总标题
%         \captionof{figure}{Analysis of accuracy with varying temporal and spatial parameters.}
%         \label{abl2}
%     \end{minipage}
% \end{wrapfigure}

\section{Conclusion}
\label{conclusion}
We propose an innovative Gaussian Aware Temporal Scaling (GATS) framework for robust spatio-temporal modeling of 4D point cloud videos, which effectively addresses distributional uncertainty and temporal scale bias through the joint design of uncertainty guided Gaussian convolution and temporal scaling attention. Distinctively, we are the first to analyze point cloud dynamics from the perspective of relative velocity, providing a principled solution to frame rate inconsistency and motion representation stability. Extensive experiments demonstrate that GATS consistently enhances performance across multiple benchmarks, highlighting both its theoretical significance and practical value for real world 4D point cloud understanding.
{
    \small
    \bibliographystyle{ieeenat_fullname}
    \bibliography{main}
}

\clearpage
\setcounter{page}{1}
\maketitlesupplementary

\section{Experimental Setup}
\subsection{Theoretical Results}

We first analyze the effect of discrete sampling on velocity estimation in point cloud videos. 
Let \(x(t): \mathbb{R} \to \mathbb{R}^3\) denote the continuous trajectory of a point in physical time. 
When the trajectory is sampled at discrete intervals \(\Delta t\), the observed sequence is $x^t = x(t_0 + t \cdot \Delta t)$. 
The corresponding discrete velocity estimator is defined as
\begin{equation}
\hat{v}(t) = \frac{x^{t+\Delta t} - x^t}{\Delta t}.
\end{equation}

This estimator deviates from the true instantaneous velocity \(v(t) = \tfrac{dx}{dt}\). 
By expanding \(x(t+\Delta t)\) around \(t\) using Taylor series, we obtain
\begin{equation}
x(t+\Delta t) = x(t) + v(t)\Delta t + \tfrac{1}{2}a(t)(\Delta t)^2 + o((\Delta t)^2),
\end{equation}
where \(a(t)\) denotes the acceleration. Substituting this into the discrete estimator yields
\begin{equation}
\hat{v}(t) = v(t) + \tfrac{1}{2}a(t)\Delta t + o(\Delta t).
\end{equation}
It follows that the estimation error \textbf{grows linearly} with \(\Delta t\), which leads to inconsistent velocity estimation across different frame rates.

To eliminate this dependency, we introduce a temporal scaling factor \(s > 0\) and define the normalized velocity estimator as
\begin{equation}
\hat{v}_s(t) = \frac{x^{t+\Delta t} - x^t}{s \cdot \Delta t}.
\end{equation}
By choosing \(s = \tfrac{\Delta t}{\Delta t_{\mathrm{ref}}}\), where \(\Delta t_{\mathrm{ref}}\) is a fixed reference interval, the estimator becomes
\begin{equation}
\hat{v}_s(t) = \frac{x(t+\Delta t) - x(t)}{\Delta t_{\mathrm{ref}}}.
\end{equation}
This normalization maps all sequences to the same reference time scale, thereby removing sampling-rate bias. Moreover, as \(\Delta t \to 0\), the normalized estimator converges to the true velocity \(v(t)\).

Next, we examine how the scaling factor behaves with respect to frame density. 
Consider a video segment of fixed physical duration \(T_{\mathrm{seg}}\) sampled into \(F\) frames. 
The discrete interval is
\begin{equation}
\Delta t = \frac{T_{\mathrm{seg}}}{F}.
\end{equation}
With a fixed reference interval \(\Delta t_{\mathrm{ref}} > 0\), the scaling factor is
\begin{equation}
s = \frac{\Delta t}{\Delta t_{\mathrm{ref}}} 
  = \frac{T_{\mathrm{seg}}}{\Delta t_{\mathrm{ref}}} \cdot \frac{1}{F}.
\end{equation}
This expression shows that \(s\) is inversely proportional to the frame count \(F\). 
Equivalently, for a fixed segment duration, increasing the number of frames (or equivalently, the frame rate) reduces the scaling factor.

Since \(\Delta t = \tfrac{T_{\mathrm{seg}}}{F}\), substituting into the definition of \(s\) yields
\begin{equation}
s(F) = \frac{T_{\mathrm{seg}}}{\Delta t_{\mathrm{ref}}} \cdot \frac{1}{F} = C \cdot F^{-1}, 
\quad C = \frac{T_{\mathrm{seg}}}{\Delta t_{\mathrm{ref}}} > 0.
\end{equation}
The function \(s(F) = C F^{-1}\) is strictly decreasing for \(F>0\), with derivative
\begin{equation}
\frac{ds}{dF} = - C F^{-2} < 0.
\end{equation}
Hence, as more frames are sampled within a fixed duration, the temporal scaling factor decreases accordingly.

Finally, if we express the relationship in terms of frame rate, let the frame rate be \(\mathrm{fps}\) and the segment length be \(T_{\mathrm{seg}}\). 
Since \(F = \mathrm{fps} \cdot T_{\mathrm{seg}}\), the scaling factor becomes
\begin{equation}
s = \frac{1}{\Delta t_{\mathrm{ref}}} \cdot \frac{1}{\mathrm{fps}}.
\end{equation}
Thus, for a fixed reference interval, an increase in frame rate leads to a reduction of the scaling factor, whereas a decrease in frame rate results in an increase of the scaling factor.
This inverse relationship ensures that the normalized velocity estimator remains invariant to frame partitioning and robust across heterogeneous video sources.

\subsection{Datasets} 
\label{sec:data}
\paragraph{MSR-Action3D} This widely used dataset, captured by a first generation Kinect sensor, comprises 567 depth sequences. It features 20 distinct human action categories, totaling approximately 23,000 frames with an average sequence length of 40 frames. To ensure a direct and fair comparison with prior art, we adhere to the conventional partition, assigning 270 sequences to the training set and 297 to the testing set.

\paragraph{NTU RGBD} As a large scale benchmark for 3D human action recognition, this dataset contains 56,880 video samples distributed across 60 fine-grained action classes. The duration of each video clip varies, ranging from 30 to 300 frames. We adopt the standard cross subject evaluation protocol, which designates 40,320 videos for training and 16,560 videos for testing.

\paragraph{Synthia 4D} To assess our model's generalization performance on dense prediction, we conduct 4D semantic segmentation experiments using the Synthia 4D dataset. Derived from the original Synthia, this dataset consists of six dynamic driving scenarios. We follow the established experimental setup from previous works, using a standard frame wise split of 19,888 frames for training, 815 for validation, and 1,886 for testing.

\subsection{Training Details} 
\label{sec:tra}
\paragraph{MSR-Action3D} We follow P4Transformer to partition the training and testing sets. For each video, we densely sample 24 frames and sample 2048 points in each frame. We train our model on a single NVIDIA A100 GPU for 50 epochs. The SGD optimizer is employed, where the initial learning rate is set as 0.01, and decays with a rate of 0.1 at the 20-th epoch and the 30-th epoch respectively.

\paragraph{NTU RGBD} Consistent with prior methodologies, our data processing involves 24-frame sequences with a temporal step of 2, sampling 2048 points per frame. The training was conducted over 15 epochs with a batch size of 24 on a single NVIDIA A100 GPU. An SGD optimizer was utilized, configured with an initial learning rate of 0.01 and a cosine decay schedule to ensure stable convergence and facilitate a fair comparison.

\paragraph{Synthia 4D} For the Synthia 4D experiments, our model was trained for a total of 150 epochs on a single NVIDIA A100 GPU, using a batch size of 8. Each input consisted of 3-frame clips, with 16,384 points per frame. The optimization was managed by an SGD optimizer with a momentum of 0.9. We implemented a learning rate schedule featuring a 10-epoch linear warmup from an initial LR of 0.01, followed by a multi-step decay (factor of 0.1) at the 50th, 80th, and 100th epochs.

\section{Network Configurations}
\paragraph{GATS Architecture} In our GATS framework, the point cloud sequence is first fed into the \textbf{Gaussian Aware Spatio-Temporal Embedding module}. In this stage, spatial neighborhoods are modeled using Gaussian weighted convolution, which leverages local mean and covariance statistics, while the temporal dimension is processed through scaled normalized convolution. This design enables the extraction of spatio-temporal feature representations that are robust to both distributional uncertainty and frame rate variations.

Subsequently, the features are passed into a \textbf{dual branch adapter (UGGC + TSA)}. The UGGC (Uncertainty Guided Gaussian Convolution) branch enhances spatial robustness by incorporating Gaussian statistics, whereas the TSA (Temporal Scaling Attention) branch introduces scaled temporal biases to ensure cross frame consistency. The outputs of these two branches are fused through an uncertainty aware gating mechanism, where the gating weights are determined by the condition number of the local covariance. 

The fused features are then processed by the \textbf{Transformer encoder}, where each layer consists of an Attention block and a FeedForward block. The Attention block extends the standard QKV projection and multi-head attention by incorporating temporal scaling bias, thereby guaranteeing invariance to frame partitioning. The FeedForward block is composed of LayerNorm, fully connected layers, and non-linear feedforward transformations, with residual connections to maintain training stability. By stacking multiple such layers, the model produces globally consistent and robust spatio-temporal representations.

\begin{figure}[ht]
    \centering
    \includegraphics[width=\linewidth]{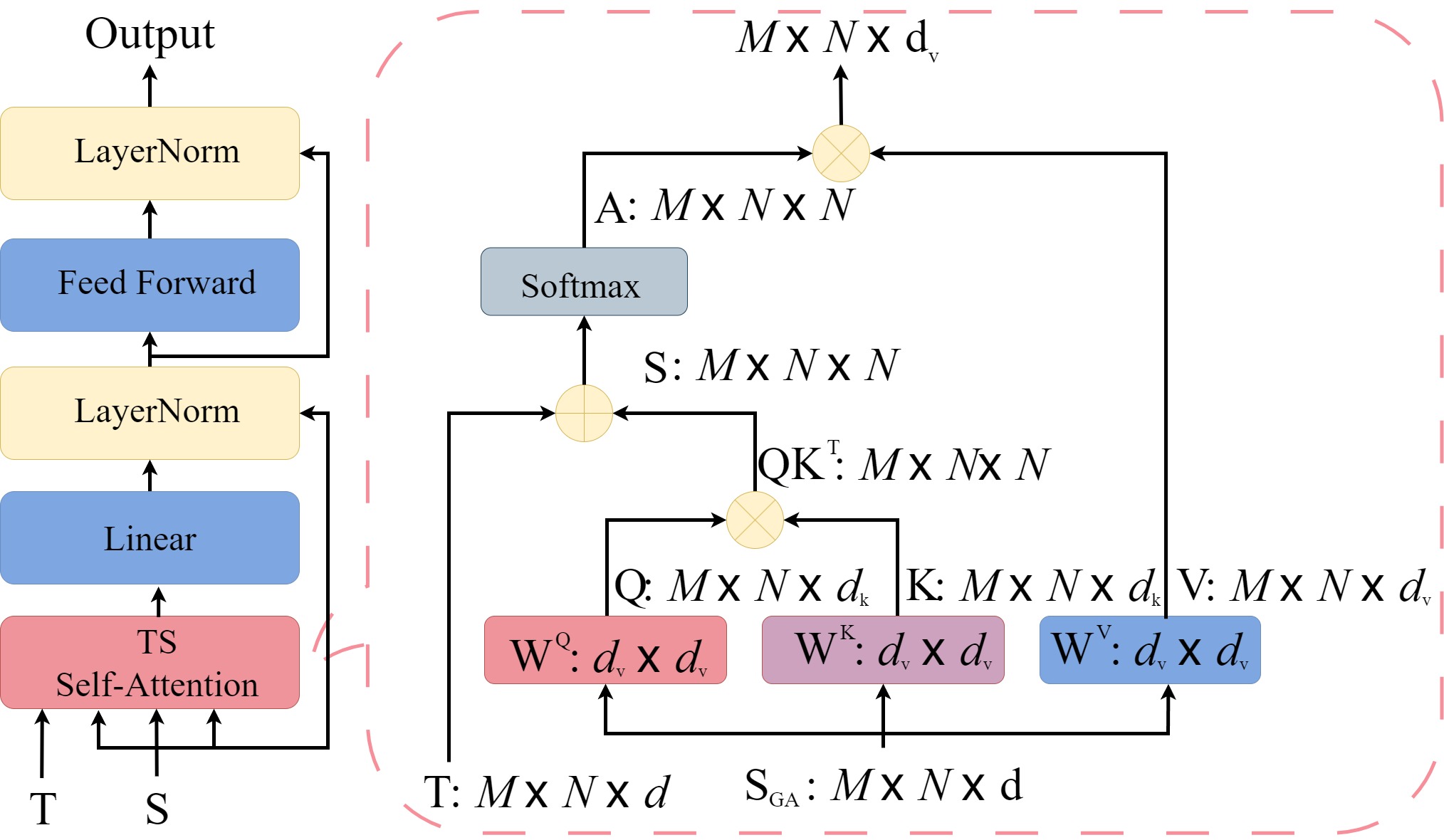}
    \caption{Structure of GATS. The left side shows the overall process, and the right side shows the
specific process of DATS.}
    \label{A1}
\end{figure}

The overall GATS architecture and the detailed attention structure are illustrated in Fig.~\ref{A1}, where \(T \in \mathbb{R}^{M \times N \times d}\) denotes the temporal tokens and \(S_{\mathrm{GA}} \in \mathbb{R}^{M \times N \times d}\) represents the Gaussian aware tokens.

\section{Additional Quantitative Results}
\paragraph{Quantitative Results} 
Table~\ref{tabA3d} expands prior comparisons, showing our method achieves 97.56\% accuracy, significantly outperforming depth-, skeleton-, and point-based baselines. Traditional depth-based methods such as Vieira et al. and Kl\"aser et al. achieve 78.20\% and 81.43\% accuracy respectively, indicating limited capacity in capturing fine-grained motion cues. Skeleton-based approaches like Actionlet improve performance to 88.21\%, benefiting from structured human pose representations. Point-based methods show mixed results: while MeteorNet reaches 88.50\%, PointNet++ lags behind at 61.61\%, highlighting the challenge of modeling spatio-temporal dynamics directly from raw point clouds. In contrast, our proposed GATS framework achieves a substantial improvement, reaching 97.56\% accuracy. This demonstrates its superior ability to capture both spatial and temporal dependencies in 4D point cloud sequences, validating its effectiveness for action recognition tasks.
\begin{table}[ht]
\caption{Action recognition accuracy comparison on a standard benchmark.}
\label{tabA3d}
\centering
% \resizebox{0.65\textwidth}{!}{
\begin{tabular}{lcr}
\toprule
\textbf{Methods} & \textbf{Input} & \textbf{Accuracy (\%)} \\
\midrule
Vieira et al. & depth & 78.20 \\
Kl\"aser et al. & depth & 81.43 \\
Actionlet & skeleton& 88.21 \\
PointNet++ & point& 61.61 \\
MeteorNet & point & 88.50 \\
\midrule
GATS(Ours) & point & \textbf{97.56} \\
\bottomrule
\end{tabular}
% }
\end{table}

\begin{figure*}[ht]
    \centering
    \includegraphics[width=0.9\textwidth]{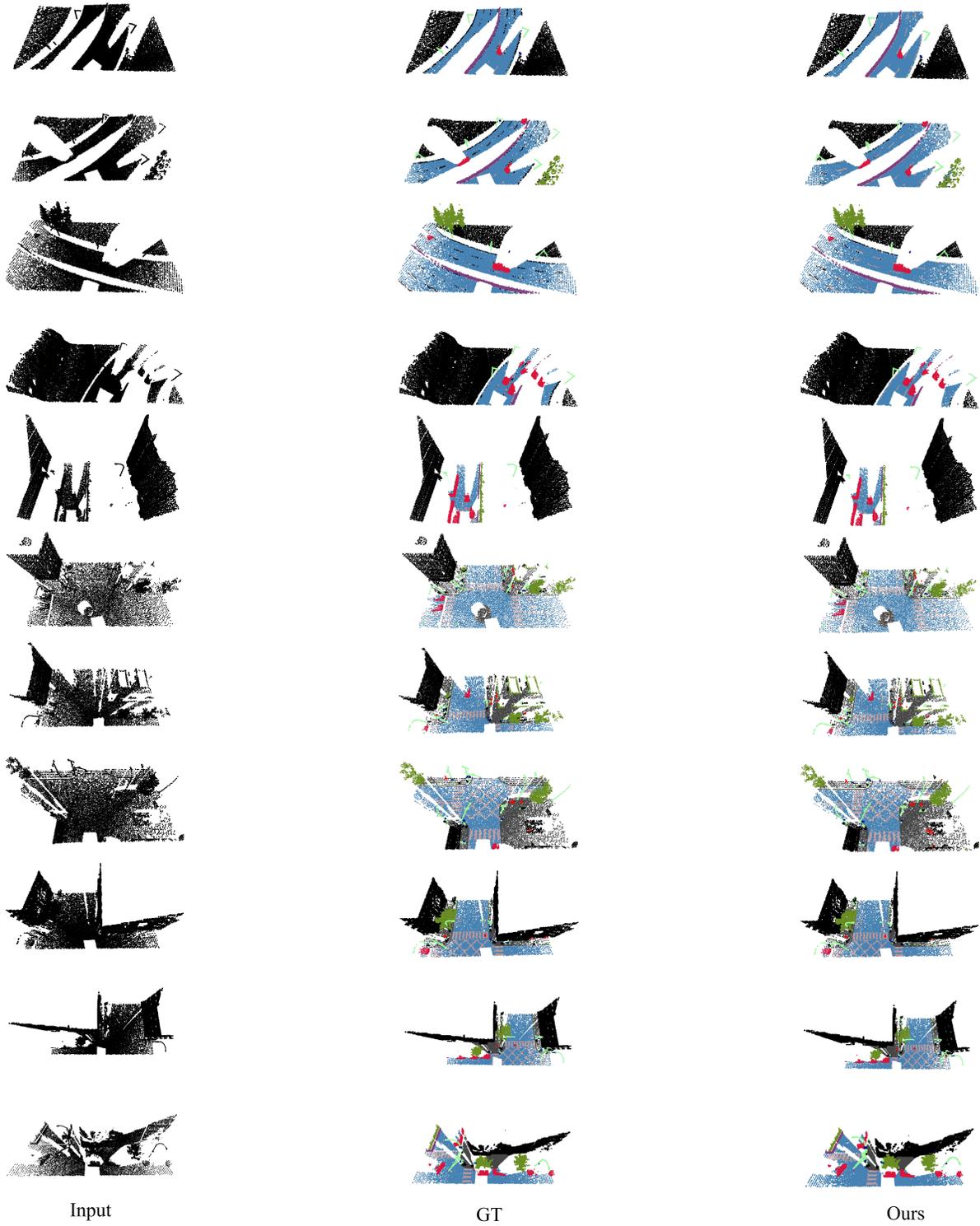}
    \caption{Qualitative results of Syn4D.}
    \label{A2}
\end{figure*}

\paragraph{Qualitative Results} Fig.~\ref{A2} demonstrates that our model consistently delivers precise segmentation across complex scenes, underscoring its robustness and generalization. The first column displays the input frames from dynamic 4D point cloud sequences, while the second column shows the corresponding ground-truth semantic labels. The third column presents the segmentation results produced by our method. As illustrated, even in scenarios with significant spatial irregularity and temporal variation, our approach maintains high accuracy in distinguishing semantic categories. These qualitative results further validate the effectiveness of our design in handling diverse and challenging environments, highlighting its strong adaptability and generalization capability.

\end{document}